**Title**: A data-driven functional projection approach for the selection of feature ranges in spectra with ICA or cluster analysis.

**Corresponding author**: Catherine Krier

**Address**: Bâtiment Maxwell
　　　　　place du Levant, 3
　　　　　B-1348 Louvain-la-Neuve
　　　　　Belgium

**Telephone**: +32 10 47 2165

**Fax**: +32.10.47.2598

**E-mail address**: krier@dice.ucl.ac.be

# A DATA-DRIVEN FUNCTIONAL PROJECTION APPROACH FOR THE SELECTION OF FEATURE RANGES IN SPECTRA WITH ICA OR CLUSTER ANALYSIS

C. KRIER[1*], F. ROSSI[3], D. FRANÇOIS[2], M. VERLEYSEN[1]

[1,2] *Université catholique de Louvain, Machine Learning Group*
[1] *DICE, Place du Levant 3, B-1348 Louvain-la-Neuve, Belgium*
*{krier, verleysen}@dice.ucl.ac.be*
[2] *CESAME, Av. G. Lemaître 4, B-1348 Louvain-la-Neuve, Belgium, francois@inma.ucl.ac.be*
[3] *Projet AxIS, INRIA-Rocquencourt, Domaine de Voluceau, Rocquencourt, B.P. 105, 78153 Le Chesnay Cedex, France,*
*Fabrice.Rossi@inria.fr*

**Abstract:** Prediction problems from spectra are largely encountered in chemometry. In addition to accurate predictions, it is often needed to extract information about which wavelengths in the spectra contribute in an effective way to the quality of the prediction. This implies to select wavelengths (or wavelength intervals), a problem associated to variable selection. In this paper, it is shown how this problem may be tackled in the specific case of smooth (for example infrared) spectra. The functional character of the spectra (their smoothness) is taken into account through a functional variable projection procedure. Contrarily to standard approaches, the projection is performed on a basis that is driven by the spectra themselves, in order to best fit their characteristics. The methodology is illustrated by two examples of functional projection, using Independent Component Analysis and functional variable clustering, respectively. The performances on two standard infrared spectra benchmarks are illustrated.

## 1. Introduction

Predicting a dependent variable from the measure of a spectrum is a frequently encountered problem in chemometrics. For example, one might try to predict the concentration in sugar, nitrogen or alcohol or the percentage of fat content, from infrared spectra measured on juice, grass, wine or meat samples, respectively. Accurate predictions may help avoiding heavy and costly chemical measurements, as spectra measures are generally much more affordable and can be automated easily.

---

[*] Corresponding author

However, accurate predictions are not sufficient in many applications. Indeed it is often required to have an insight on which chemical components are responsible for the prediction, in order to interpret the model and further develop models and products. As basic chemical components are associated to limited spectra ranges, finding which parts (ranges) of the spectra are responsible for the prediction is important in many applications too.

For this reason, prediction models that use only parts of the spectra (after variable selection) will be preferred, when interpretation is sought. Not all prediction models are able to force the fact that only parts of the spectrum are used. For example, the traditional PLS uses loadings that are combinations of all spectral variables. Even if some of them have a larger weight in the model than others, in most situations it can hardly be concluded that only some parts of the spectra participate to the prediction model.

Selecting ranges in spectra may be seen as a variable selection problem where the variables are the wavelengths in the spectrum. The variable selection problem at hand is specific in the sense that its dimensionality is high. The dimension is indeed the number $N$ of wavelengths available in spectra, which can easily reach several hundreds or thousands in modern, high-resolution spectrometers. Obviously, with such high resolution, consecutive wavelengths are highly correlated. This leads to the well-known colinearity problem, and all other difficulties related to the high dimension and the dependencies between coordinates (complex models, difficulties of convergence in the latter, overfitting, curse of dimensionality, etc.).

In addition, selecting individual variables is not a prefect solution though: because of the colinearity between consecutive variables, there is often no more reason to select an individual wavelength than the preceding or next one. The selection of groups of consecutive variables (or wavelength ranges) seems thus less arbitrary and can in addition reduce the computational burden of the selection process. Selecting groups of consecutive wavelengths may be achieved through the so-called functional modeling approach: contrarily to vector approaches (which are by far the most common ones in spectrometric modeling), the order (or indexes) of variables is taken into account. In other words, the information contained in the fact that an infrared spectrum is a smooth function of the wavelengths is not lost. Remind that in vector approaches, such as PLS, the numbering of variables is irrelevant: any swapping or permutation will rigorously lead to the same prediction results; the smoothness of spectra, although an important information, is not exploited.

The principle of functional approaches is to project the spectra on a basis of smooth functions, and then to use the coordinates of each spectrum on the basis as new variables [1]. In this way, the number of variables is reduced in the projection, and the functional nature of spectra is exploited through the smoothness of the basis functions. For example, B-splines are often used as basis functions [2], [3], as they offer several advantages: the projection is easy to compute, the basis functions cover approximately equal ranges, and taking into account the derivative of spectra is easily achieved [4]. Projection on wavelets [5], [6] also preserves the functional nature of spectra, but do not have the equal range property.

In both cases however, the basis is chosen a priori, without taking the spectra into account. In other words, the same basis is used for any problem (an exception is the adjustment of some parameters, such as the number of B-splines). Obviously, this solution might not be optimal: "better" bases could be found if the spectra themselves were taken into account. For example, it may happen that some parts of the spectra are highly informative, necessitating an increased number of basis functions in that part, while others are flat and/or not correlated to the dependent variable, thus necessitating less or no basis functions. It is thus preferable to choose the basis, or the groups of consecutive wavelengths, in a spectra-driven way, for increased prediction performances and interpretation.

Selecting appropriate groups of wavelengths according to the specificities of the spectra is the idea developed in [7]. In this paper, an information-theoretic criterion is used to group together (to cluster) variables that carry similar information content. As the clustering is driven by the value of the spectra, the size of the groups is automatically adjusted to optimize the prediction performances. However, the approach in [7] is not functional, in the sense that the indexes of the spectral variables are not used. The procedure may result in the selection of consecutive variables, as a consequence of their colinearity, but it may not be the case too.

In this paper, we develop a framework for the functional analysis of smooth spectra, through their projection on bases that are spectra-driven. In that sense, we combine the advantages of functional analysis (taking a priori into account the smooth nature of spectra) and of data-driven variable grouping (using a basis adapted to the data, for increased prediction performances and interpretation). Two examples of this data-driven functional approach are detailed. The first one decomposes the spectra into independent components that will be further used as basis functions. Independent Component Analysis (ICA) [8] is used for the decomposition. Because of the independence between basis functions, it will be shown that they occupy a limited range of the spectral variables, providing interpretability. The second one borrows the variable clustering idea from [7], replacing a group of variables by a "mean" characteristic, though exploiting the functional nature of the data.

The two proposed approaches are intended to select ranges of spectrometric variables, and not only isolated variables. They differ in the fact that the clustering methodology is a crisp method, in the sense that each spectral variable is assigned to a single cluster, while this is not exactly the case for the ICA approach, where each spectral variable has a certain "weight" in each cluster (but close to zero for most of them). It is expected that selecting ranges of spectral variables will help interpretability, in particular to identify basic components that are contained in the samples measured by spectrometry. This does not mean however that the goal of the method is to extract the spectra themselves of the basic component; this would be a much more ambitious goal, and would require further independence hypotheses between components that are not found in practice.

This paper is organized as follows. Section 2 reviews recent methods to select variables in spectrometric modeling. The proposed methodology is then presented in Section 3 at a general level. Details relating to the two projection basis envisaged in this paper are given in the next section. Section 5 explains how to select the projected variables in order to build a prediction model. Finally, Section 6 presents and discusses the experimental results on two Infra-Red spectroscopy datasets

## 2. Theory: state-of-the-art

This section presents state-of-the-art methods to select variables in spectrometric modeling. The first part is about functional approaches with a priori fixed bases, and the second part addresses a variable clustering, but non-functional, solution.

### 2.1. *Functional projection*

In general, the Infra-Red (IR) spectra can be seen as smooth functions sampled at different wavelengths corresponding to the spectral variables. This allows considering functional analysis as a tool to model and interpret such data.

B-splines are commonly used as a basis to project the spectra into a lower dimensional space [2]. These functions present indeed a "local" behavior, in the sense that each spline corresponds to a specific range of spectral variables; this specificity preserves the interpretability of the results. Moreover, the groups of spectral variables defined by the splines contain only consecutives spectral variables, which present often a high degree of colinearity. Splines, and thus ranges of spectral variables, can then be selected through their coefficients. This selection is often based on the mutual information between the coefficients and the parameter to predict [9].

The choice of this kind of functional basis is however made a priori, without taking the information included in the spectral variables into account, which is certainly not optimal. This means also that different problems will be handled with the same functional basis. In addition, all ranges of wavelengths defined by the splines share the same size, and are not adapted to the shape (peaks and flats) of the spectra.

### 2.2. *Clustering*

Another way of grouping similar spectral variables is the clustering. Van Dijk and Van Hulle propose in [7] to use a hierarchical clustering with a similarity notion adapted to spectral variables and clusters of variables. This similarity is based on the mutual information between the spectral variables.

Contrarily to the functional projection, clusters can have different sizes, adapted to the particularities of the spectra. The information included in the spectral variables is also exploited when determining the clustering in the similarity measure. This clustering is however not a functional projection approach, because spectral variables grouped in a cluster are not necessarily consecutive in the spectra; two variables corresponding to different wavelengths can indeed be grouped in a cluster, while another variable related to an intermediate wavelength is included in a second one. This phenomenon, avoided by the projection on B-splines, may be problematic from the interpretability point of view.

## 3. Methodology

In the present paper, the prediction problem is solved in a three-step procedure. First, spectra are replaced by their projection on a functional basis. Two types of bases are considered here: a basis formed by the independent components (computed by ICA) of the spectra, and clusters of spectral variables. A classical preprocessing such as derivation, Standard Normal Variate (SNV) [10] or Multiplicative Scatter Correction (MSC)[11] , can be applied before the projection.

In the first case the new variables are the coefficients of the projection on the independent components. In the second case the new variables are mean values taken over the clusters. The projection of the spectra is unsupervised for both the ICA and clustering approaches (only the number of clusters in this second case is chosen in a supervised manner), in the sense that the projection is not designed explicitly to optimize the quality of the prediction of the dependent variable.

Second, the new variables are selected by a supervised criterion. Indeed the number of variables resulting from the first step, which makes no or little use of the dependent variable, will generally be too large for an efficient use in the prediction model. Therefore, the second step consists in selecting which of the resulting variables will be used in the prediction model.

The proposed feature selection methods are linear in the sense that the final selected features are specific linear combination of the original features, both in the case of the ICA and the case of variable clustering. A linear model built on those variables is also a linear model build on the original variables and is thus likely to have lower performances than a PLSR model which is optimal among such linear models. Consequently, a nonlinear model has to be considered in order to preserve the quality of the prediction.

As a nonlinear prediction model will be used, the criterion used to select the variables must be nonlinear too. The Mutual Information (MI) will be used, as a nonparametric nonlinear criterion to evaluate the usefulness of variables, in a greedy procedure (forward search followed by an exhaustive search).

Third, the prediction model itself will be built on the variables selected after step two. The nonlinear model used in the experimental section is a traditional Least-Squares Support Vector Machines (LS-SVM) [12]. Other models such as Multi-Layer Perceptrons (MLP), Support Vector Machines (SVM) and Radial-Basis Function Network (RBFN) [13] could be used too.

Sections 4 and 5 will detail the functional projection and variables selection steps, respectively. Section 6 will present some experimental results.

## 4. Functional projection

This section shows two possible ways to reduce the number of variables in spectra. The functional character is preserved in both cases. The first solution consists in projecting the spectra on a basis found by Independent Component Analysis. The second solution consists in clustering the variables under a functional constraint.

**4.1. *ICA***

Independent Component Analysis (ICA) [8] is a signal processing technique aiming at separating measured signals into their "sources" components. Sources are considered as the basis components which are mixed to produce the measured signals. The general problem of recovering sources from the measured signals is called Blind Source Separation (BSS). It is blind because the mixture process (from the sources to the measured signals) is unknown and unobserved. To be solved, this problem required further hypotheses. The first one is that the sources are statistically independent; this leads to the so-called Independent Component Analysis (ICA). It is also necessary to make some hypotheses on the mixture. Though other options are possible, the hypothesis that will be made here is that the mixing process is linear and instantaneous. Let us define $S$ the *kxn* matrix of source signals $s_i$, **Z** the *pxn* matrix of observed signals $z_i$, and **A** the *p×k* mixing matrix, where *k* is the number of independent sources, *p* the number of observed signals, and *n* the number of observations per signal. Then the model may be written as

$$\mathbf{Z} = \mathbf{A\,S}.$$

The ICA problem reduces in finding estimations $\hat{\mathbf{S}}$ of the sources by combining the observed signals **Z**:

$$\hat{\mathbf{S}} = \mathbf{WZ} = \mathbf{WA\,S}.$$

Numerous methods based on some measure of the independence between the estimations $\hat{\mathbf{S}}$ are able to find an appropriate matrix **W**. Note that the ICA solution suffers from two indeterminacies, a possible permutation and a scaling of the sources. The resulting **WA** product will therefore not necessarily be the identity matrix; it will be a matrix with one non-zero element per line and per column.

In our context of infrared spectra analysis, the observations $z_i$ are the spectra of the analyzed products, and the sources $s_i$ are supposed to be the spectra of their independent basic chemical components. The implicit reasonable hypothesis of the above model is thus that the spectrum of a mixing of basic components is a linear combination (with coefficients corresponding to the proportions) of the component spectra.

In the above model each column of **A** corresponds to a single source. The column is thus considered as the set of coefficients (for each analyzed spectrum) corresponding to this source, in other words the coefficients of the projection on this source. The second step mentioned in Section 3 and detailed in Section 5 will therefore consist in selecting columns of **A**.

A last question remains to be discussed in the context of the ICA projection: the choice of the number of independent components. Because of the intrinsic formulation of the ICA problem, this number cannot exceed the number of observed spectra. However, a lower number is preferred, both to facilitate the subsequent variable selection, and to avoid numerical problems in the ICA solution. A simple solution to the choice of the number of independent components is to reconstruct the spectra (from **Z** = **A S**) with a limited number of sources, and to choose the lower number which leads to a small reconstruction error (the threshold may be fixed according to the level of noise in the data).

**4.2. *Clustering***

The spectral variable clustering method described in this section is based on the combination of the variable clustering algorithm proposed in [7] with functional constraints that facilitate the interpretation of clusters of variables.

When a smooth function is discretized over a fine grid, the resulting values are generally very similar on sub-intervals of the original range with "reasonable size": smoothness implies that $f(w_1)$ and $f(w_2)$ are very close when $w_1$ and $w_2$ are. As a consequence, spectral variables corresponding to close wavelengths take nearly identical values. It is therefore tempting to reduce the resolution of the spectra by down-sampling them over wavelength ranges in which they are almost constant (on each individual spectrum). This can be done by clustering identical or similar variables together, and then by replacing each cluster of variables by the average of the variables it contains. This procedure is similar to a standard *k*-means algorithm, applied here to variables (features) instead of observations. The prototype summarizing a cluster is therefore a vector whose number of elements is the number of spectra: it corresponds to a new variable.

However, the goal of this procedure is to reduce the number of features while maintaining interpretation possibilities. It is therefore important to design the clustering procedure such that it produces clusters that fulfill the latter requirement. This is done first by requesting clusters to contain only consecutive variables: each cluster is associated this way to a wavelength range that is much easier to interpret than a set of arbitrary and possibly unrelated spectral variables. It should be noted that this requirement is not artificial: while a spectral variable is naturally more similar to a spectral variable with a close wavelength, the very smooth nature of some spectra leads to "long distance" similarities, as illustrated on Figure 1. This Figure represents the absolute value of the correlation between a spectral variable and all the other variables for the Tecator dataset (see Section 6 for details on this dataset). While the first three maximal correlations correspond to the variable itself and its two neighbors (leading to the wavelength range [906.6, 910.6]), the fourth maximally correlated variable corresponds to the wavelength 942.9 nm. Clustering variables without functional constraint could thus lead to disconnected hardly interpretable clusters.

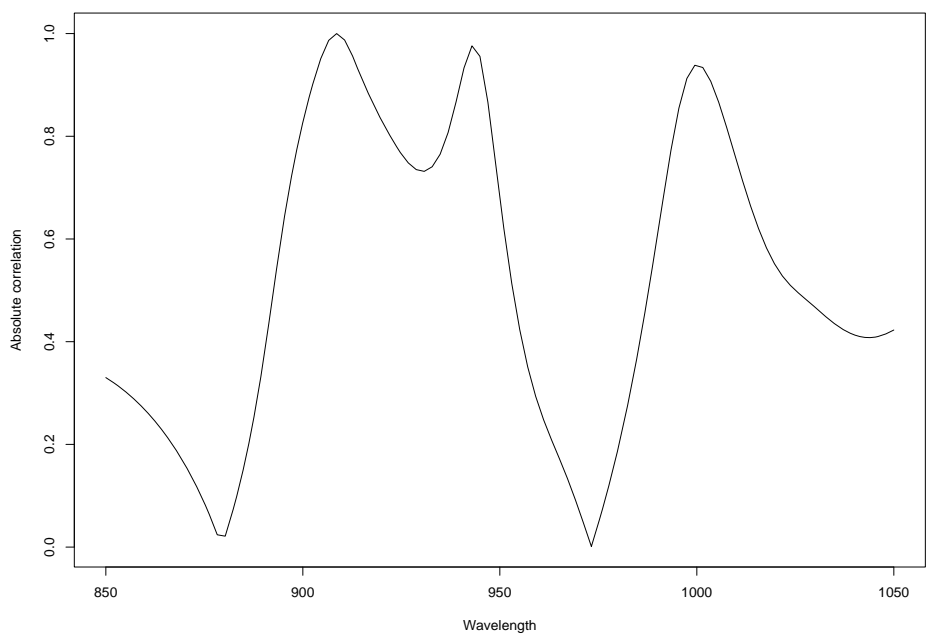

Figure 1: Absolute Correlation between a spectral variable and all the other variables for the Tecator dataset.

The second element used to enforce easy interpretation is the use of a simple similarity criterion between variables, namely the absolute correlation, *i.e.*

$$\Sigma(X_i, X_j) = \left| \frac{E(X_i X_j) - E(X_i)E(X_j)}{\sqrt{Var(X_i)Var(X_j)}} \right|.$$

This measure has several advantages over the mutual information used in [7]. It is both easier and faster to evaluate via standard estimators. However, the most important aspect is that the correlation measures only simple (linear) dependencies between variables, whereas the mutual information evaluates complex nonlinear relationships. When it is used to build a predictive model, this is a valuable characteristic, but in the case of variables clustering, it can lead to clusters that are quite difficult to interpret. For instance, if $X_2=(X_1)^2$, the mutual information between the variables will be high; they will tend to be considered as similar and hence put in the same cluster. However, the actual values of $X_2$ and $X_1$ will be quite different, a confusing fact for the experts. Moreover, the possible complex nonlinear relationship between variables in a cluster turns the design of a synthesis (for example mean) variable into a difficult problem. If variables in a cluster are highly correlated, they are also highly correlated to the mean variable of the cluster, while if they have high mutual information, they have no particular reason to have a high mutual information with the mean variable. Indeed, a high mutual information between two variables corresponds to the existence of an approximate bijective function mapping one variable to the other: even if there are exact one-to-one mappings between several variables in a cluster, there might be no such mapping between those variables and their mean.

In practice, the proposed method is based on an agglomerative (bottom-up) full linkage hierarchical clustering algorithm. It proceeds as a standard hierarchical clustering method, with the additional constraints of merging only adjacent clusters. For $N$ original spectral variables, the algorithm works as follows:

1. Create $N$ clusters, $c_1=\{X_1\}$, ..., $c_N=\{X_N\}$.

2. Compute the similarity between each cluster and its following cluster, *i.e.*, between $c_i$ and $c_{i+1}$, as the absolute correlation between the variables they contain (*i.e.*, $X_i$ and $X_{i+1}$).

3. Repeat *N*-1 times:

    a. Find the pair of *consecutive* clusters that are the most similar.

    b. Merge the clusters.

    c. Update the similarities between consecutive clusters as follows: the similarity between $c_i$ and $c_{i+1}$ is the minimum of the absolute correlation between any pair of variables, the first one being in $c_i$ and the second in $c_{i+1}$.

At each step of the algorithm, the number of clusters is reduced by one, starting from *N* and going down to 1. Clusters have a hierarchical structure, summarized by a dendrogram. For each number of clusters between 1 and *N*, the original variables are replaced by one variable per cluster, defined as the mean of the variables contained in the cluster. As clusters correspond to sub-interval of the original wavelength range, the resulting new variables can be considered as forming a piecewise constant approximation of the original spectra. In this sense, the proposed methodology is related to the B-spline based functional representation used in [9] and recalled in Section 2. The major difference is that [9] used B-splines of order 4 or 5, with regularly spaced knots. The piecewise constant approximation build here corresponds to B-splines of order 1, with non-regular spacing between knots.

The algorithm described above constructs iteratively *N* successive clusterings (*N-2* non-trivial ones). In order to choose one of them, the expected performances of the complete model (variable clustering + variable selection + nonlinear prediction) would be an ideal selection criterion. In theory this is achievable. In practice however, a robust evaluation of each nonlinear model would require computationally-demanding techniques such as cross-validation, leading to unaffordable computation times when $N$ is large. An approach that does not make use of the full model is thus preferred. A linear measure (the correlation) between features being used to build the hierarchical clustering, it is suggested to use a linear model to evaluate each of the $N$ clustering results. More precisely, for each $M$ between 1 and $N$, the cross-validated performances of a linear regression model built on the clustering result with $M$ classes are calculated. The optimal $M$ is chosen as the one that maximizes this performance indicator.

## 5. Feature selection

New variables are extracted from the spectra after a projection or clustering procedure as detailed in the previous section. However, in the projection case the output to predict (the dependent variable) is not taken into account. In the clustering approach, it is not fully taken into account either: the similarity criterion used to group the original features together makes no use of the output to predict, the latter being used only to choose among the $N$ clustering results. However, in this case, only a linear predictive model is used. Therefore, in both cases, it is expected that the result of the projection or clustering step will contain a prohibitive number of variables (although reduced compared to the number of original variables in the spectra) to be entered directly in a nonlinear prediction model. Indeed the projection or clustering step groups together features that are sufficiently correlated, but does not provide a way to eliminate those (or the resulting clusters) that carry no information to predict the output. A further feature selection may thus be considered, at least if the selection criterion used for this step makes use both of the output variable, and of the nonlinear potential of the prediction model; the goal here is to select only the variables that are relevant for the prediction.

A selection procedure consists in two basic blocs: the relevance criterion and the selection procedure. The solution described in the following sections is similar to the ones used in our previous works [9], [14].

## 5.1. *Relevance criterion*

First, a relevance criterion has to be chosen: how to measure if a variable, or a group of variables, is pertinent for the prediction (in the present context, a group of variables is not anymore a group of spectral variables, but rather a group of new variables obtained after functional processing). The best choice for the criterion is certainly to estimate the performances of the model itself. However, let us remind that we are in a context where the number $M$ of variables is high. Testing all $2^M-1$ models will obviously lead to a prohibitive computational cost. In addition with a greedy (forward) procedure as detailed below, the model performance criterion is not adequate. Indeed the forward procedure will begin with a single variable, then two, etc., and there is no reason that the model itself will perform well on a too small number of variables. Instead of an approach based on prediction performances, a simpler one is preferred, where the criterion is evaluated independently from the model. The correlation could have been used, but it suffers from two drawbacks: it measures only linear relations, and is most usually limited to individual variables, not groups of. Therefore, the Mutual Information (MI) is preferred. The MI between **X** and **Y** is defined as

$$MI(\mathbf{X},\mathbf{Y}) = \int \mu_{X,Y}(x,y) \log \frac{\mu_{X,Y}(x,y)}{\mu_X(x)\mu_Y(y)} dxdy \tag{1}$$

where $\mu_X$, $\mu_Y$ and $\mu_{XY}$ define the probability density function of variables **X**, **Y** and the joint **X**,**Y** variable respectively. Note that **X** and **Y** in the above definition may be multidimensional, which answers to the above mentioned requirement to handle groups of variables. In our context, **X** will denote one or a group of variables obtained after projection or clustering, and **Y** is the dependent variable.

The mutual information must of course be estimated on the learning set. The evaluation of the mutual information by equation (1) requires the knowledge of the densities. In practice, their estimation on the learning set is unreliable when the number of variables in **X** increases, due to the curse of dimensionality. A better estimator has been proposed in [15] to handle multi-dimensional variables. This estimator is based on the search for nearest neighbors among the learning set in the spaces spanned by **X** and **Y**, therefore avoiding the direct estimation of the densities; it will be used in the experimental section of this paper.

### 5.2. *Forward selection*

Once the criterion is chosen, a procedure to select variables has to be defined. Remind that the variables considered here are those resulting from the first step, *i.e.* the projection or feature clustering. Each of them represents a cluster of initial variables that are consecutive in the spectra. The further selection described in this section does not force to select clusters that are consecutive. The idea is that several ranges of spectral variables may be of interest for the prediction problem (see Section 6 for examples); it is thus important to let the method select non-consecutive clusters.

As explained above, an exhaustive search on all combinations of variables among the set of available ones would be computationally prohibitive. A forward selection is preferred. It consists in iteratively selecting variables at each step of the procedure, without questioning a choice previously made. The first selected variable is the one which maximizes the MI with the dependent variable **Y**. The second one is chosen as to maximize the MI between **Y** and the group of two variables formed by the previously selected one and the new one to choose. The procedure is repeated by adding one variable at each step.

In practice, the estimated value of the MI might decrease at some step of the forward selection. This is however only a consequence of the bias and variance of the estimator. In theory indeed, the true value of the MI can only increase when variables are added in **X**. Therefore the maximum of MI cannot be used as stopping criterion (although such choice in sometimes found in scientific publications). Rather, the forward procedure is iterated until a "reasonable" number of variables is selected. By "reasonable", we mean a number $P$ for which 1) the curse of dimensionality remains limited in the MI estimator, and 2) the following exhaustive search remains computationally affordable. In practice, the value of $P$ may for example be limited to 7 or 8.

Finally, an exhaustive search on all remaining variables is performed. In Section 5.1 it was argued that building $2^M$-1 non-linear models would have been prohibitive when $M$ is large. Now than $M$ is replaced by $P$, where $P$ equals for example 7 or 8, building $2^P$-1 models becomes affordable, even when nonlinear models are considered. Therefore $2^P$-1 nonlinear models are built, their performance evaluated (for example by cross-validation), and the one which leads to a minimum of the estimated error is selected. In the next Section about experiments, LS-SVM models are used for this purpose. The details of the learning procedure for these models are given in [4].

In practice, the forward selection proceeds as follows:

1. Define $V_1$ as the $X_i$ that maximizes *MI($X_i$;* **Y***)*.
2. For $k$ from 2 to *P*,
    - define $V_k$ as the $X_i$ that maximizes MI($V_1,\ldots,V_{k-1}, X_i$);**Y**).

3. For each subset $L$ of $\{V_1,\ldots,V_P\}$,
   - build a LS-SVM on $L$,
   - evaluate the best performances of the LS-SVM by cross-validation.
4. Select the best subset as the one that maximizes the cross-validation performances.

## 6. Experiments

This Section introduces first the datasets from the food industry used in the paper to evaluate the proposed methods. The next part details the experimental methodology. The results are then presented and discussed.

### 6.1. *Datasets*

Two datasets are used to illustrate and evaluate the functional approaches described in this paper. The first database is Tecator [16]. It consists of 215 near-infrared spectra of meat samples, recorded on a Tecator Infratec Food and Feed Analyzer in the 850–1050 nm wavelength range. Each spectrum is discretized into 100 spectral variables. The spectral variables are the absorbance, defined by $\log(1/T)$, where $T$ is the measured transmittance. All spectra have been normalized according to the SNV method (mean equal to zero and variance equal to 1). The spectra are used to predict the fat content of the meat samples and are divided into a learning and a test set. The learning set defined for the experiments contains 172 spectra, which leaves 43 spectra for the test.

The second dataset, Wine [17], is related to the alcohol concentration in wine samples. In this case, spectra consist of absorbance measures recorded in the mid-infrared range at 256 different wavenumbers between 4000 and 400 $cm^{-1}$; the number of spectral variables is therefore 256. The learning and test sets contain respectively 94 and 30 spectra. Three spectra (numbers 34, 35 and 84) are however considered as outliers and consequently removed from the learning set.

**6.2.** *Experimental methodology*

The experimental methodology is summarized in Figure 2. For each database, the functional approaches described in Sections 4 and 5 are compared to two more "classical" approaches in similar experimental conditions: the fist one is a non-linear model (LS-SVM) and the second one is a standard PLSR. These experiments are conducted on the original data to provide additional reference performances.

All experiments are carried out on identical training and test sets. In order to choose the meta-parameters for the LS-SVM (*i.e.* the regularization parameter and standard deviations of the kernel), a 3-fold cross-validation procedure is carried out. Again, the same subsets of spectra are used for all compared methodologies. These different approaches are implemented in Matlab® version 7.1, using the LS-SVM Toolbox (LS-SVMLab1.5) [18]. In the case PLSR, processing has been done with the R statistical software [19], using the pls package [20].

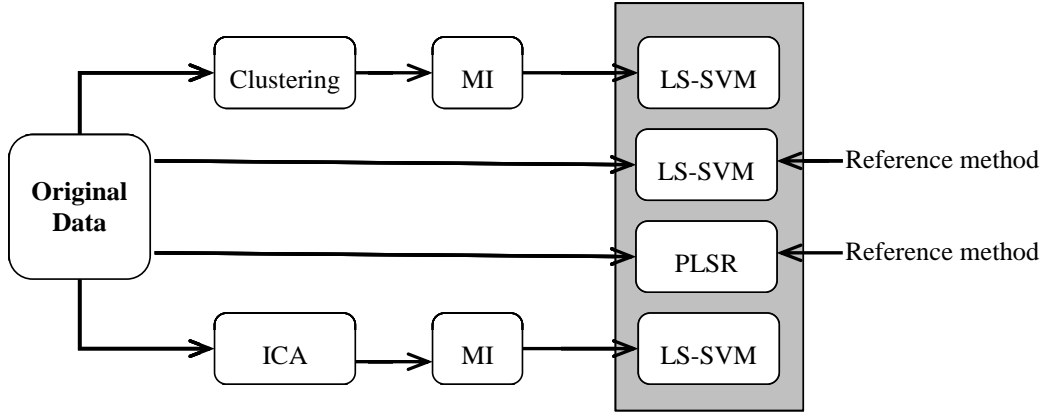

Figure 2: Summary of data processing methods.

In the case of the two proposed approaches, the spectra are projected on the functional basis for the two datasets. The projected variables (respectively the independent components and the representatives of the clusters) are then selected by mutual information and a LS-SVM is built on the selected projected variables.

The ICA is performed by the *FastICA* version 2.5 toolbox [21]. The options chosen for the analysis are: "symmetric" (estimate all the independent components in parallel) and "stabilization" (stabilized version of FatsICA). The non-linear function used for the experiments is $g(u) = u^3$. The clustering procedure leads to 27 clusters for Tecator and 30 for Wine. Concerning the selection of the projected variables by MI, the forward procedure is stopped at 7 variables, which means $2^7-1$ LS-SVM models built on all possible combinations of variables. As suggested in [15], the mutual information estimator takes 6 neighbors into account; this neighborhood is defined according to the Euclidian norm.

The performances of the predictive models are evaluated on a test set $\Omega$, by the Normalized Mean Squared Error (NMSE). The NMSE is defined as follows:

$$NMSE_\Omega = \frac{\frac{1}{Q_\Omega} \sum_\Omega (\hat{Y}_i - Y_i)^2}{Var(Y)},$$

where $Y_i$ is the parameter to predict, $\hat{Y}_i$ the corresponding prediction and $Q_\Omega$ the number of data in $\Omega$. The NMSE is thus always non-negative; a NMSE is greater than 1 means that the prediction performances are worst than if all predicted values where equal to the mean of $Y$ [22] (actually, the NMSE is equal to $\frac{Q_\Omega}{Q_\Omega - 1}$) in this extreme case.

The most time-consuming step is the forward selection carried out after the clustering or the projection of the spectra on the independent components; these last steps are very fast (a few second). The whole procedure takes between 30 minutes and an hour.

**6.3. *Results***

This section presents the results of each considered methodology in terms of prediction performances, for the two datasets.

6.3.1. *Tecator*

The ICA leads to a decomposition of 12 independent components for the Tecator dataset. In the case of this dataset, it must be mentioned that the number of the independent components has been limited to 12, because large instabilities in the results of the ICA algorithm were observed when it was tried to extract more independent components. However with 12 independent components, the reconstruction error of the spectra was around 35%, which is not good. This might be seen as a severe limitation in this case: to reach around 1% of reconstruction error, around 30 independent components would have been needed, inducing the instability problems. Nevertheless, it must be mentioned that even if the reconstruction error of the spectra is large, good prediction results might be obtained. Indeed, it might happen that the lost information in the reconstruction is not relevant for the prediction. In other words, for example, the prediction model will use only small ranges in the spectra, while completely discarding other parts that consequently do not have to be reconstructed correctly. It is therefore important to check, a posteriori, if the quantitative results obtained with the method are comparable (even if not equal or better) to those obtained by methods using all variables, regardless of the reconstruction error. This will be done below in Table 1.

Among the 172 spectra used for learning, two subsets of 57 spectra and a subset of 58 spectra are used for the 3-fold cross-validation procedure. Table 1 presents the NMSE on the test set (43 spectra) for the three considered approaches, as well as the number of variables involved in the LS-SVM (in the case of ICA and clustering, this means the number of independent components and clusters, respectively). The best prediction performances are achieved by the clustering methodology, closely followed by the "classical" model. The ICA approach give acceptable but worse results, while PLSR gives the worst results. It should be noted that all results, even the worst ones, correspond to good prediction performances. This means that interpreting the selected variables should be meaningful. In both case the functional projection allows one to reduce the number of variables significantly.

Figure 3 shows in grey the ranges of wavelengths selected by the clustering methodology; the vertical lines correspond to the limits of the clusters. The selected independent components provided by the ICA approach are indicated in Figure 4. As expected, they contain a limited range of spectral variables, while of course, as detailed in the introduction of this paper, they do not strictly correspond to pure spectra of pure components.

| Method | Number of (latent) variables | NMSE test |
|---|---|---|
| PLSR on raw data | 11 | 0.0170 |
| LS-SVM | 100 | 0.0086 |
| ICA + MI + LS-SVM | 7 | 0.0135 |
| **Clustering + MI + LS-SVM** | **5** | **0.0077** |

Table 1: Prediction performances for Tecator.

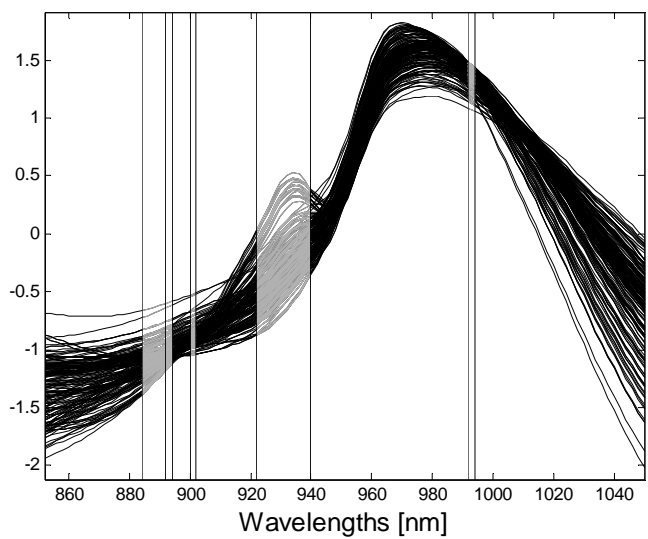

Figure 3: Tecator spectra and spectral variables selected by the clustering approach (in grey).

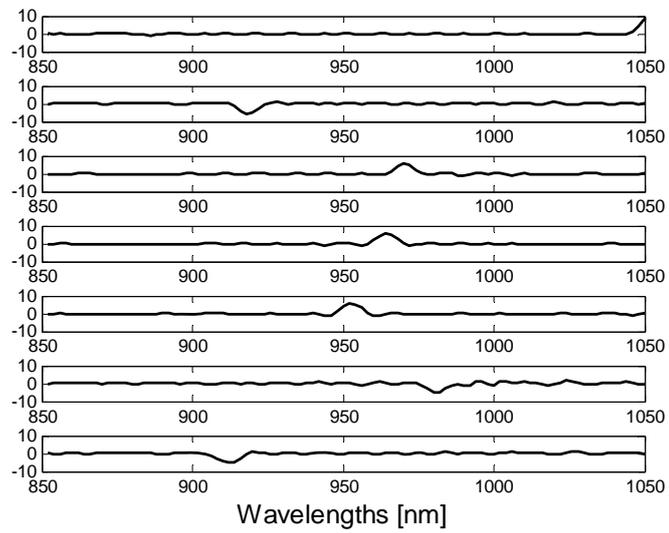

Figure 4: Selected independent components for Tecator.

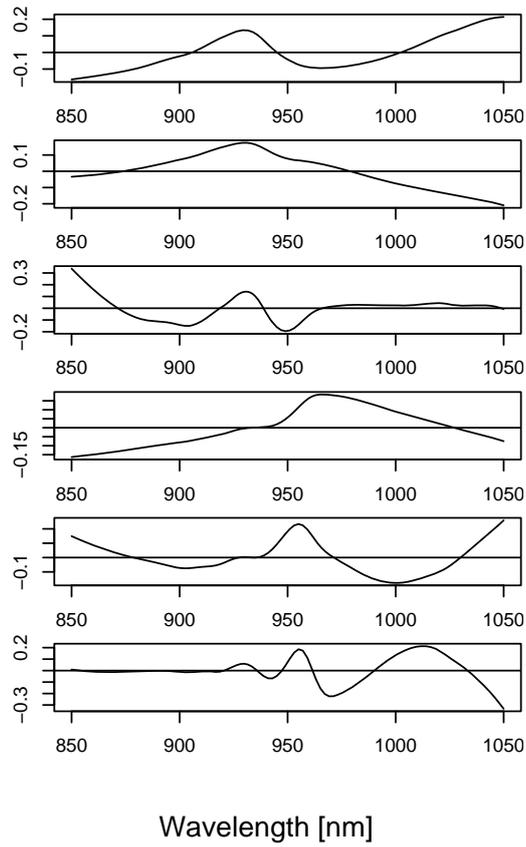

Figure 5: The six first selected loadings for PLSR in the case of Tecator.

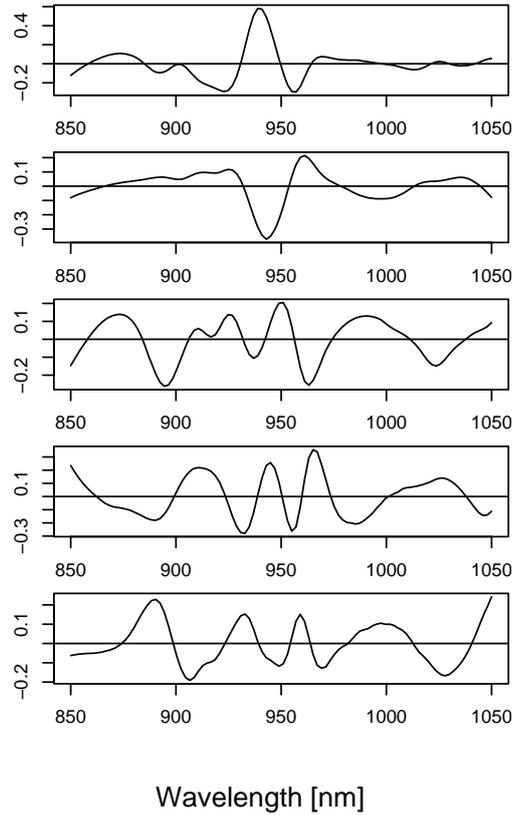

Figure 6: The five last selected loadings for PLSR in the case of Tecator.

The two proposed approaches select wavelengths around 930 nm, which corresponds to a bump in the shape of the spectra. This result is in agreement with the results e.g. in [14]. In addition, the large size of the cluster covering this wavelengths range tends to show that the whole bump in the shape of the spectra plays the same role in the prediction of the meat samples fat content. Both ICA and clustering methodologies select wavelengths in the neighborhood of 880-910 nm, which is also in conformity with what can be found in the literature. However, the first independent component is quite different from the wavelengths usually selected. This is probably due to the change of the baseline around 1040 nm.

As a comparison, Figures 5 and 6 give the loadings of the PLSR. While they show some interesting trends, they are obviously far more difficult to interpret than the ICA components (not to mention the wavelength cluster). This is mainly due to the fact that most of the coefficients of the loadings are non zero. There are of course some wavelength intervals with zero influence on some specific loadings (e.g., [975, 1050] for the third loading on Figure 5), but none of them is as focused as ICA components or wavelength clusters.

On the Tecator dataset, the functional approaches lead to good prediction quality based on localized wavelength ranges, something that can not be achieved by PLSR. Similar results were obtained in [14] by selecting individual spectral variables but at the expense of a much higher computation cost (roughly 500 time more processing time, *i.e.* more than 10 day of calculation compared to half an hour on identical hardware).

6.3.2. *Wine*

In the case of the Wine database, the ICA leads to 30 independent components. For this example, the number of independent components has been chosen only according to a reconstruction error argument, as no problem of ICA instability was observed. With 30 components, the reconstruction error goes below 1%, and does not significantly reduce if more components are used. On the contrary, using less than 30 independent components makes the reconstruction error increase significantly (10% already with 27 components).

Among the 91 spectra used for learning, two subsets of 30 spectra and one subset of 31 spectra are used for the 3-fold cross-validation procedure. The NMSE on the test set (30 spectra) and the number of variables used by the LS-SVM are detailed in Table 2, together with the result of PLSR. The lines of matrix **A** are normalized (reduction and centering) before building the LS-SVM. The best performances in prediction are obtained by the LS-SVM model build on all spectral variables, followed by PLSR and by the clustering approach. Those three methods give good prediction performances. As for the Tecator dataset, this means that the wavenumber ranges selected by the clustering approach should be meaningful. On the contrary, the ICA approach has rather bad performances: this gives an indication that care must be exercised when interpreting ICA components in this case. As for the Tecator dataset, the functional approach reduces the number of variables used by the nonlinear model.

Figure 7 indicates in grey the ranges of wavenumbers selected by the clustering methodology. The vertical lines delimitate the clusters. Figure 8 shows the selected independent component. As expected, the wavenumber ranges selected by the clustering approach are very satisfactory in the context of alcohol concentration prediction: they correspond to the absorption range of the O-H bond present in alcohol (around 3600 $cm^{-1}$). While PLS loadings lead to similar conclusion (see Figures 9 and 10), this is far less clear in their case. Indeed most of the loadings associate high coefficients to wavenumbers around 3600 $cm^{-1}$, but they also give an as large weight to wavenumbers around 1500 $cm^{-1}$. More generally, while the situation is better for PLS than in the case of the Tecator dataset, loadings remain spread over the whole spectral range and are therefore more difficult to interpret than cluster of wavenumbers.

| Method | Number of (latent) variables | NMSE test |
|---|---|---|
| PLSR on raw data | 8 | 0.0058 |
| LS-SVM | 256 | 0.0042 |
| ICA + MI + LS-SVM | 5 | 0.3777 |
| **Clustering + MI + LS-SVM** | **3** | **0.0086** |

Table 2: Prediction performances for Wine.

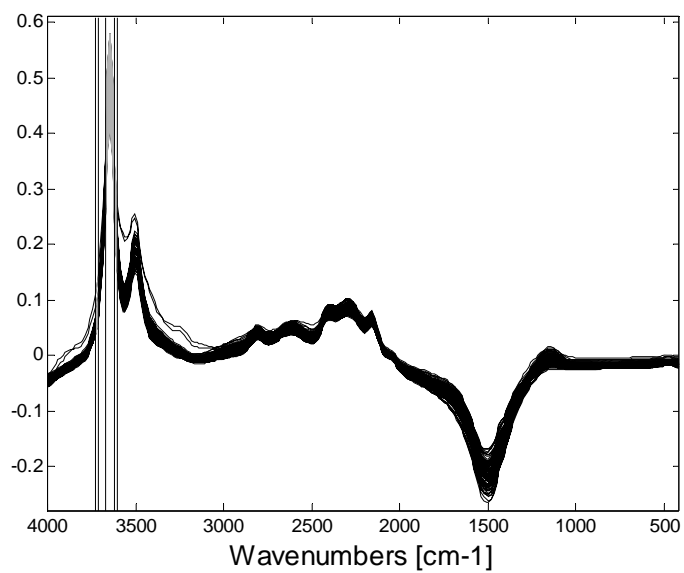

Figure 7: Wine spectra and spectral variables selected by the clustering approach (in grey).

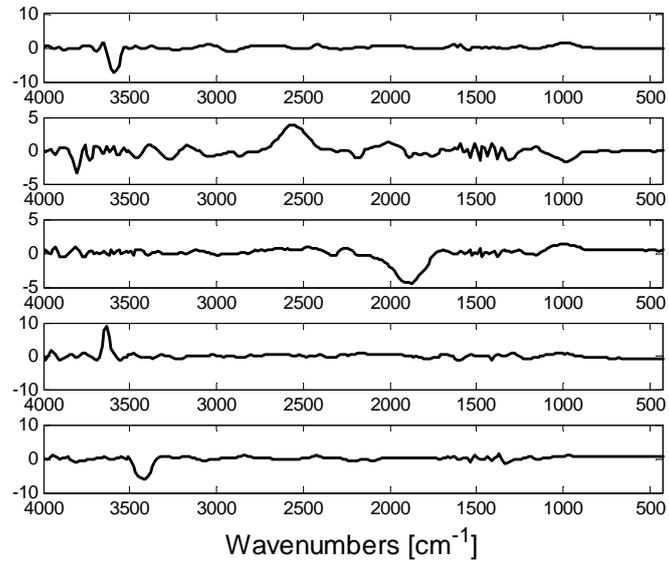

Figure 8: Selected independent components for Wine.

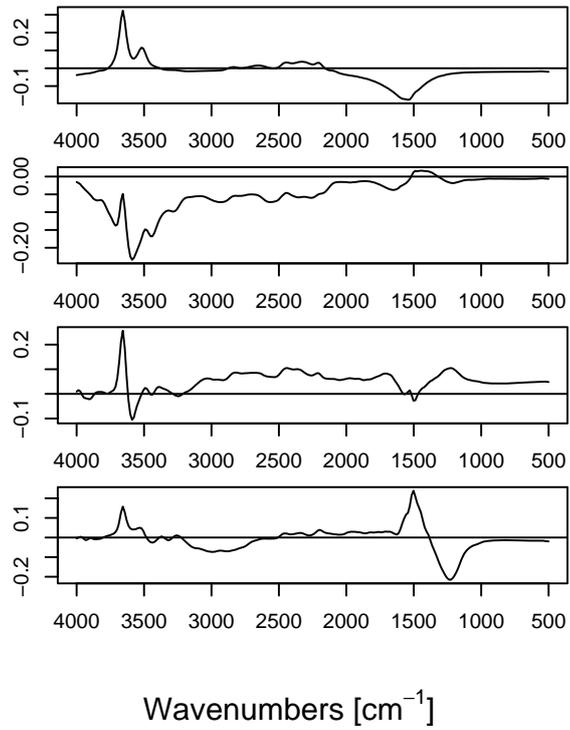

Figure 9: The four first selected loadings for PLSR in the case of Wine.

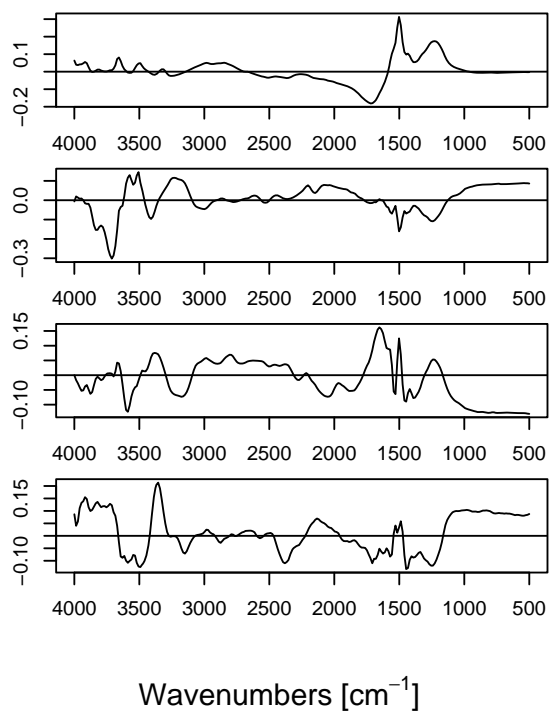

Figure 10: The four last selected loadings for PLSR in the case of Wine.

The case of ICA is less clear as its predictive performances are not very good. Nevertheless, from the five selected components, only the third one does not correspond to a part of the spectra which is selected by the clustering approach, while the second component is a mixed between some wavenumbers around 3800 cm$^{-1}$ and a large bump centered on 2600 cm$^{-1}$. Moreover, parts of the spectra defined by the second and fifth independent components are close to wavenumbers selected in [23], [24], in which they gave good predictive performances. Therefore, despite bad predictive performances, the components remain associated to interesting part of the spectra. A possible explanation of the differences between the clustering approach and ICA on this dataset is the very thin aspect of the cluster retained by the first method (which is a consequence of the high number of clusters originally retained by the linear selection method). As it is partly supervised, it can detect that a rather high precision is needed in some parts of the spectra in order to achieve good predictive performances. On the contrary, ICA components appear to have a rather large support (compared to the clusters) which are then incompatible with high quality prediction. This explanation is partly confirmed by the spiky aspect of the PLS loadings. On the contrary, the Tecator dataset corresponds to smoother loadings and to wider variable clusters. As already mentioned, this illustrates the need to compare the results to state-of-the-art methods using all variables (and not looking for interpretability).

## 7. Conclusion

Reducing the number of variables taken into account in a prediction model for spectrometric data is not only important to avoid problems related to the curse of dimensionality (such as overfitting or difficulties of convergence), but also helps to increase the interpretability. Identifying the parts of the spectra that play a role is indeed a crucial issue for the practitioners.

Existing methods projecting the spectra into a lower-dimensional space present several drawbacks: the projection of the spectra on a basis such as B-splines is not adapted to the shape of the spectra and does not exploit the information on the parameter to predict that can be found in the spectral variables. In contrast, the clustering approach developed by Van Dijk and Van Hulle in [7] is data-driven, but does not exploit the functional character of the spectra. As a result, clusters are allowed to contain non-consecutive spectral variables, which may be a problem from the interpretability point of view.

In this paper, we suggest a methodology to address the variable projection problem while avoiding the drawbacks pinpointed above. This approach is based on a functional clustering or the projection of the spectra on independent components. The projected variables are then selected according to the mutual information between them and the parameter of interest.

Both methods result in limited ranges of spectra used for the prediction model. These ranges may be exploited by practitioners to understand on a chemical point of view which pure components might be mostly responsible for the performances of the prediction model. It is not expected however to retrieve the spectra themselves of the pure components, as this would require further independence hypotheses that are not met in practice.

It must be mentioned that the goal of the methods presented in this paper is qualitative, in the sense that finding limited ranges of spectral variables is the objective. Quantitatively, the performances of the models may vary considerably. This is due to the fact that the procedure is designed as a compromise between computational burden and performances. In some situations, it may occur that state-of-the-art prediction performances would require more variables, or more clusters than affordable by the method. When using the methods presented in this paper, it is therefore of good practice to compare their quantitative results (prediction performances) with state-of-the-art (linear and nonlinear) methods that do not look for interpretability. The closest the performances are from the latter, the best confidence can be put in the quality of the resulting interpretation.

Experiments on the Tecator and Wine databases show that the models built on the projected variables can give accurate predictions. The projections lead to a significant reduction of the number of variables to take into account in the models. Moreover, the selected wavelengths ranges correspond mainly to wavelengths identified as meaningful in the literature.

Clustering and ICA are given here as examples of functional preprocessing (projection), aimed at considerably reducing the number of variables used for the prediction, while exploiting the functional character of spectra for increasing the interpretability of the results. Further work will consider other types of functional projection, and how to choose between the different projection possibilities to obtain a good compromise between prediction performances and interpretability.

**Acknowledgments**

The authors thank the anonymous referees for their valuable suggestions that helped to improve this paper. C. Krier is funded by a Belgian FRIA grant. Parts of this research results from the Belgian Program on Interuniversity Attraction Poles, initiated by the Belgian Federal Science Policy Office. The scientific responsibility rests with its authors.